\documentclass{article} % For LaTeX2e
\usepackage[preprint]{colm2026_conference}
\usepackage{microtype}
\usepackage{hyperref}
\usepackage{url}
\usepackage{booktabs}
\usepackage{listings}

\usepackage{adjustbox}
\usepackage{xspace}
\usepackage{amsmath}
\usepackage{lineno}

\definecolor{darkblue}{rgb}{0, 0, 0.5}
\hypersetup{colorlinks=true, citecolor=darkblue, linkcolor=darkblue, urlcolor=darkblue}
\usepackage{tikz}
\usetikzlibrary{positioning, arrows.meta}

\def\sysname{{\textsc{Lift}}\xspace}
\def\lift{{\textsc{Lift}}\xspace}
\def\etoe{{\textsc{Eeft}}\xspace}
\def\sd{{\textsc{Sd}}\xspace}

\newcommand\ignore[1]{{}}

\title{\sysname: Last-Mile Fine-Tuning for Table Explicitation}

\author{Divij Khaitan \\
Microsoft Corporation\\
Bangalore, India \\
\texttt{\{t-dkhaitan\}@microsoft.com} \\
\And
Ashish Tiwari\\
Microsoft Corporation \\
Redmond, WA 98052 USA \\
\texttt{\{ashish.tiwari\}@microsoft.com}
}

\begin{document}

\ifcolmsubmission
\linenumbers
\fi

\maketitle

\begin{abstract}
    We propose last-mile fine-tuning, or \lift, a pipeline in which a pre-trained large language model extracts an initial table from unstructured clipboard text, and a fine-tuned small language model (1B–24B parameters SLM) repairs errors in the extracted table. On a benchmark of 2,596 tables from three datasets, \lift matches or exceeds end-to-end SLM fine-tuning on tree-edit-distance-based similarity (TEDS) metric while requiring as little as 1,000 training examples — where it outperforms end-to-end fine-tuning by up to 0.144 TEDS points. We term this approach last-mile fine-tuning and show it also more robust to input format variability.
    Comparisons with self-debug and end-to-end fine-tuning approaches show that last-mile fine-tuning provides an
attractive option when training data is limited or when robustness to input variation is sought without compromising on accuracy.
\end{abstract}

\section{Introduction}
    Tables are a convenient means for structuring large amounts of data to make it interpretable to humans. Having data in tabular format enables analyses of the data using table-processing software, such as Microsoft Excel. However, very often tables are shared using PDF documents, such as, in academic papers or financial reports. We envision a copy-paste paradigm for extracting tables from PDF documents; that is, a user selects the table (or a part of a table) of interest, copies it, and then pastes it {\em{as a table}}. Apart from its simplicity and ease of use, the copy-paste paradigm also has other benefits: it gives control to the user on what they want to extract as a table. Furthermore, it also eliminates any requirement to share or upload the full PDF document to extract tables. 

    When a user copies a table from a PDF document, the clipboard is populated with some unstructured text containing the data in the table and almost negligible information about the table structure. 
    Extracting a table from this unstructured text is a problem that has not been extensively studied, but there is some very recent work~\citep{TEN}, which serves as our inspiration.

    Extracting tables from PDF documents is challenging. A PDF document may be image-based (scanned PDFs) or text-based. 
    Image-based PDFs require techniques such as optical character recognition (OCR) and are out-of-scope for us since it is 
    not even possible to {\em{select}} regions from such a document to {{copy}}. 
    When PDFs are text-based, it is possible to select regions and copy a table, or parts of a table, from it. 
    What shows up in clipboard depends on the PDF reader being used by the user. 
    For example, Acrobat readers can put content in rich-text format (RTF) in the clipboard in some cases.
    We do not make any assumptions on the reader the user is using, and hence work with clipboard contents
    dumped by the most basic PDF readers, where the clipboard just contains the textual data in the selected region 
    without much information about the table topology, as in~\citet{TEN}.

    When considering table extraction from PDFs, a common paradigm is where the user uploads the PDF to a server and the software
    extracts tables while having access to the full PDF document. This is typically done using symbolic heuristics~\citep{tabula,adobe}.
    The quality of output is better when the software performs native PDF structural analysis, which is possible for certain
    text-based PDFs and when one has access to code that implements the PDF specification~\citep{adobe}. When that is not the case,
    then table extraction is not accurate and almost always require the human to repair the generated table~\citep{tabula}. 

    Large Language Models (LLM) have shown proficiency on several text-based tasks, such as document summarization and sentiment analysis, even
    when they are not explicitly trained for those tasks~\citep{chowdhery2022palmscalinglanguagemodeling, zhang2022optopenpretrainedtransformer, zhang-etal-2024-benchmarking}. Such emergent behaviors in LLMs through in-context learning have turned LLMs into general-purpose transformers, though they are noisy and probabilistic in nature. It is, thus, natural to apply a LLM to extract tables from PDF documents. Again, several LLM-based agents provide the option of uploading a PDF document and then they can extract tables when prompted to do so. 

    Our interest is in extracting tables from the clipboard text generated when a user copies a selected region from a text-based PDF document. This is more difficult problem than extracting tables from PDF since we do not have access to the document here. Additionally, this is a more general problem and potentially more widely useful because it makes fewer assumptions about what we start with. In fact it can also potentially apply when the starting text is not necessarily from a PDF document, but maybe from a text email or an application log. We shall follow the lead of~\citet{TEN} and call this the {\em{table explicitation problem}}.

    A good baseline is to use pre-trained LLMs to perform table explicitation using in-context learning. This is also the first step of our last-mile fine-tuning approach. After this first step, there are two options. One option would be to do an iterative loop, in self-debug style, to fix any errors in the table generated in the previous iteration~\citep{TEN}. In this paper, we explore the second option where we repair the generated table using a small language model (SLM) specifically fine-tuned for the repair task. Small Language Models (SLM) have shown the capacity to display comparable performance to LLMs through fine-tuning on specific tasks, while offering significant benefits in terms of performance and cost of inference~\citep{ayala2025finetuneslmpromptllm, gu2024minillmknowledgedistillationlarge}. Apart from the self-debug and last-mile fine-tuning approaches, there is another option of fine-tuning an SLM for the end-to-end table explicitation task. We compare these approaches on a table explicitation benchmark that we built for this purpose.

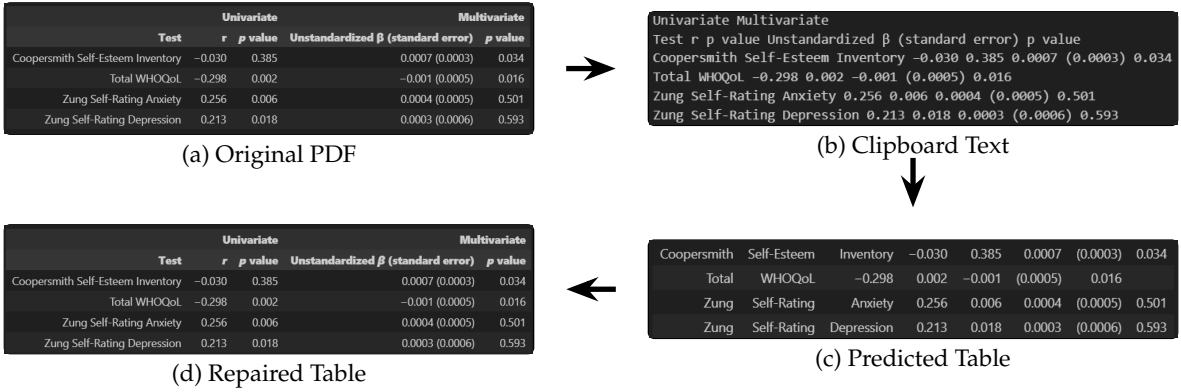
\begin{figure}[t]
\centering
    \makebox[\textwidth][c]{
    \begin{tikzpicture}[
        % Define node distance: vertical then horizontal
        node distance=1.5cm and 1.5cm,
        % Style for image containers
        img_node/.style={
            inner sep=0pt, % Removes space between image and border
            draw,          % Adds a border around the image
            line width=1pt,
            rounded corners=2pt
        },
        label_node/.style={below, node distance=5pt, font=\small},
        arrow/.style={
            -{Stealth[scale=1]},
            line width=2pt,
            shorten >= 12pt,
            shorten <= 12pt
        }
    ]

    % Place the nodes with images inside the braces {}
    % Note: Change the filenames to your actual local image files
    \node [img_node] (a) {\includegraphics[width=0.5\textwidth]{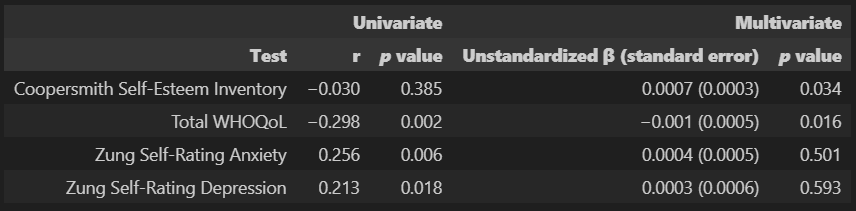}};
        \node[label_node] at (a.south) {(a) Original PDF};
    
    \node [img_node, right=of a] (b) {\includegraphics[width=0.5\textwidth]{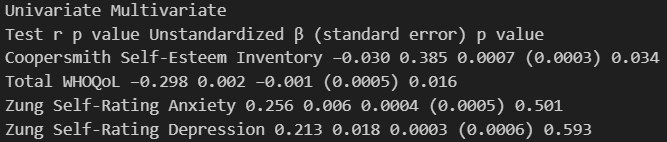}};
        \node[label_node] at (b.south) {(b) Clipboard Text};
    
    \node [img_node, below=of b] (c) {\includegraphics[width=0.5\textwidth]{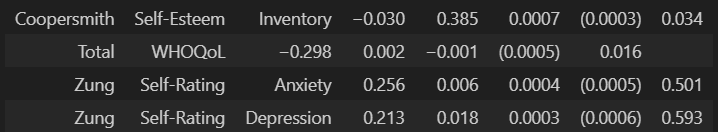}};
        \node[label_node] at (c.south) {(c) Predicted Table};
    
    \node [img_node, left=of c] (d) {\includegraphics[width=0.5\textwidth]{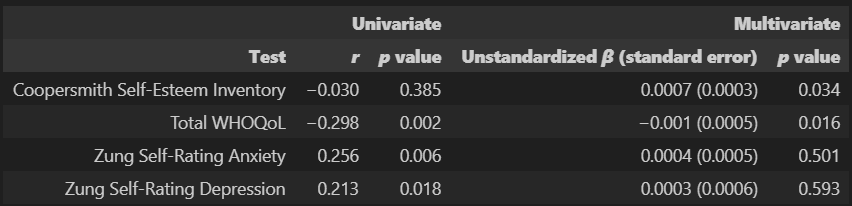}};
        \node[label_node] at (d.south) {(d) Repaired Table};
    
    % Draw the arrows connecting the node names (a, b, c, d)
    \draw [arrow] (a) -- (b);
    \draw [arrow] (b) -- (c);
    \draw [arrow] (c) -- (d);
    
    \end{tikzpicture}
    }
    \caption{{\small{Illustrating last-mile repair on table explicitation problem. The user selects a table
    from a PDF document (Figure~(a)) and copies it. The text copied into the clipboard is shown in Figure~(b).
    A baseline method (symbolic or using an LLM) is used to extract table from this text, but it generates a
    wrong table (Figure~(c)). The table is repaired using an SLM fine-tuned for last-mile repair to give us
    back the original table (Figure~(d)).}}}\label{fig:overview}
\end{figure}
            
\subsection{Overview}

Let us consider a concrete example of the table explicitation problem. 
Figure~\ref{fig:overview} illustrates the full workflow of last-mile fine-tuning approach.
Imagine here a user working on a PDF document that contains a table shown in 
Figure~\ref{fig:overview}(a). Say the user wants to copy this table to a table processing application.
The user selects the table and copies it. Our task is now to paste the copied content as a tabular
structure. Unfortunately, the clipboard often only contains the raw text containing the contents
of the table, but no structural information. This text is shown in Figure~\ref{fig:overview}(b).
Turning this text into a structured table is the {\em{table explicitation}} problem. 
There are several approaches here. The last-mile fine-tuning approach, or \lift in short, has two steps. In  the
first step, we
attempt to solve the problem using some baseline method, which can be a heuristic rule-based
method or an LLM-based approach.  Since the source text has very little structural information,
what we get after the first step is a ``broken'' table, such as the one shown in
Figure~\ref{fig:overview}(c). In the second step, this table is repaired. We fine-tune a small
language model (SLM) for just this last-mile repair task. The fine-tuned SLM is used in the second
step to repair the broken table and finally yield the correct table, shown in Figure~\ref{fig:overview}(d).

        \subsection{Related Work}
Related work can be divided into two parts: work related to last-mile fine-tuning and work related to table explicitation.

Related to last-mile fine-tuning, a similar two-stage approach has been proposed for code generation in low-resource languages. 
Specifically,~\citet{lamirage} proposed fixing an LLM-generated broken program using a neurosymbolic approach (but without any
fine-tuning). In subsequent work, the repair phase was enhanced with program repair techniques implemented using LLMs~\citep{RING}.
Fine-tuning for the repair task, which is central to \lift, appears in some recent work~\citep{SLMFix}, but again it was for 
code repair. 
Based on similar intuitions as \lift, the work Plug-in and Fine-tune~\citep{PiFi} proposes an architecture where a frozen LLM is embedded in a tunable SLM and the combination is fine-tuned for end-to-end Natural Language understanding and generation tasks. Although it is end-to-end fine-tuning, the work~\citep{PiFi} has flavors of \lift and shows good generalization capabilities. 
We apply \lift to table explicitation and, unlike earlier works, we identify trade-offs
between the three approaches, self-debug, \lift, and end-to-end fine-tuning. We observe that all three approaches
are viable options depending on the application constraints. Furthermore, in conjunction with the work on \lift for code generation, 
our work also establishes that \lift is a general principle that applies to  multiple tasks.
% : an immutable LLM generates code, a static validator flags errors, and a fine-tuned 500M SLM repairs them, achieving >95% validator pass rates. 
% LaMirage (Bavishi et al., OOPSLA 2022) formalized the "last-mile repair" concept for low-code formula languages using neurosymbolic techniques, combining symbolic candidate generation with neural ranking. 

Related to table explicitation, recent work~\citep{TEN} uses a self-debug loop, supported with a symbolic checker for this problem. A different approach based on end-to-end fine-tuning has also been used for several table-related tasks (not table explicitation)~\citep{TableLLMSpecialist}.  Table extraction has been extensively explored from various different sources.
        Extracting tables from images is one popular field of study, where the table extraction task is further subdivided into two subtasks:
            \begin{itemize}
                \item 
                    Table Structure Recognition (TSR), which refers to the problem of extracting the structural layout of a table from an image, usually in the form of a markup language or some simplified version of it, without extracting the content of individual cells. TableNet \cite{8978013} uses an encoder backbone from VGG-19, with two separate decoders to extract the main tex segments and the different column segments respectively. \citet{zhang2024semv2tableseparationline} introduces a dataset of real-world images for the TSR task, and proposes a method for detection based on instance segmentation using CNNs and spliting-then-merging. GNNs \citep{li2020gftegraphbasedfinancialtable}, transformers \citep{yang2022tableformerrobusttransformermodeling} and VLMS\citep{chen-etal-2023-tablevlm} have also been used for this task. 
                \item 
                    Table Detection, which refers to the problem of identifying a bounding box around a table from an image. This is often the first step in a larger pipeline for structural recognition or extraction, such as \citet{Prasad_2020_CVPR_Workshops}. 
                % \item 
                %     Table Recognition: identifying the structure and contents of a table from an image end-to-end, with the output in a markup language such as \LaTeX or HTML. This is usually a combination of both of the above tasks, with an OCR pipeline or a PDF text extractor to populate the table with the extracted structure.  
            \end{itemize}
            Tabularis Revilio~\citep{10.1145/3627673.3680000}  puts forth a three step pipeline for the table explicitation task, starting with a dedicated header detection module, then using an LLM to generate a sketch of the table given the header and raw text and finally using a symbolic extractor on the raw text, guided by the generated sketch to rank candidates. Our work could be applied as a post-processor to this method as well. \citet{2021arXiv210902707W} introduced the {\em{text-to-table}} task, which is the extraction of data from a long piece of text into a tabular format. It is expressed as the inverse problem of the table-to-text, where a language model is asked to generate descriptions of information presented in the table, usually in response to a query. \cite{deng-etal-2024-text} proposes a multi step approach, that extracts structured tuples from a dataset and postprocesses those tuples into a table. Our work differs from these in that rather than being a given a long piece of natural language text, the clipboard text we start with just contains contents of the table. The starting text in text-to-table has more context and also more irrelevant information.
            Table explicitation  is fundamentally different from the text-to-table task. %Additionally, the table-to-text task can have  information corresponding to different tables intertwined with each that needs to be formatted into separate tables, with the language model needing to make that distinction. Our table explicitation task expects only a single table (that can be divided into related subtables) as output, with that table's content forming most of the input text with only minimal extraneous text.

\subsection{Contributions}
We make three contributions. First, we show that replacing the 
iterative LLM self-debug loop of~\citet{TEN} with a single pass through a fine-tuned SLM 
achieves comparable or superior accuracy at significantly lower inference cost (one LLM call + one SLM call vs. two LLM calls). 
Second, we demonstrate that \lift requires less training data than end-to-end fine-tuning: with only 1,000 examples, \lift with Mistral-7B achieves 0.875 tree similarity score versus 0.731 for end-to-end fine-tuning. Third, we show \lift is more robust to out-of-distribution input formats than end-to-end fine-tuning, inheriting the LLM's ability to generalize to unseen formats.
%We revisit the table explicitation problem and evaluate the last-mile fine-tuning approach on it.
%In particular, we propose fine-tuning SLMs for last-mile repair of table explicitation tasks. 
%We show that last-mile fine-tuning is more effective than self-debug (where the base non-fine-tuned model is
%used for repair), and is competitive with end-to-end fine-tuned models.
%We also highlight two benefits of the last-mile fine-tuning approach.
%First, we show that last-mile fine-tuning is an effective strategy when there is limited data (and resources) 
%available for fine-tuning compared to the end-to-end fine-tuning approach. Second, we show that last-mile fine-tuning is more robust to
%inputs that are different from what may be in some initial training set.
While our experiments focus on table explicitation, the \lift framework is task-agnostic: any task where an LLM produces a noisy initial output that can be systematically repaired is a candidate. Recent work on code repair using small models supports this hypothesis, though direct empirical validation on other tasks remains future work.
%Combining our observations with the past work on last-mile repair for code generation (in low resource languages), it becomes clear that the last-mile fine-tuning approach is a general principle that is potentially applicable on a wider range of tasks. 
For such tasks, our work shows that  \lift is a promising approach to use when training data is scarce and when resiliency is desired.

%[TODO: Decide on user-interaction part - include, say what? Add results.]

%[TODO: ADD A motivating example.]
            
\section{Last-Mile Fine-Tuning}
\label{sec:LIFT}

Consider the task of transforming an input $x\in X$ to an output $y\in Y$ such that
some requirement $\Phi(x,y)$ holds between the input-output pair $(x,y)$.
Due to their ability for in-context learning, a pre-trained LLM can be prompted to solve
this task to give us a function $f$ that predicts an output $\hat{y} = f(x)$ for input $x$.
If $\Phi(x,\hat{y})$ evaluates to $\mathtt{True}$, then we can return the prediction $\hat{y}$.
However, if $\Phi(x,\hat{y})$ is $\mathtt{False}$, then the output $\hat{y}$ needs to be 
{\em{repaired}}.
The {\em{last-mile repair task}} is the problem of transforming the predicted output
$\hat{y}$ to the output $y$ such that $\Phi(x,y)$ holds.

The last-mile repair task can again be solved by suitably prompting a pre-trained LLM.
This approach is usually called the {\em{self-debug}} approach.
Let $f'$ denote the repair function obtained by prompting an LLM to generate $y$ 
given $x$ and $\hat{y}$. The self-debug approach
returns $\hat{y} = f'(x, f(x))$ given input $x$. An alternate to self-debug is the {\em{last-mile
fine-tuning}}, or \sysname, approach where a SLM is fine-tuned on the task of predicting
$y$ given $x$ and $\hat{y}$ task. Let $f_s$ denote
the function computed by the fine-tuned SLM. Thus, the last-mile fine-tuning approach returns
$f_s(x, f(x))$ given the input $x$. We assume only one iteration of repair here, but the repair
function can be applied multiple times guided by the check $\Phi$ in both approaches.

The probability, $Pr_{\sysname}(y \mid x)$, of generating $y$ given $x$ in the 
\sysname-approach can be computed as follows: 
$$Pr_{\sysname}(y \mid x) = \sum_{\hat{y}} Pr_{LLM}(\hat{y}\mid x) Pr_{SLM}(y \mid x, \hat{y})$$
This probability will be higher than $Pr_{LLM}(y\mid x)$ if the repair module has high probability
of correctly repairing some incorrect $\hat{y}$ answers.

A third approach here is the end-to-end fine-tuning approach, or \etoe, where an SLM is
fine-tuned for the task of predicting $y$ given $x$. If $f_{sf}$ is the function computed
by the end-to-end fine-tuned SLM, then the \etoe approach return $f_{sf}(x)$ given the input $x$.

\section{Table Explicitation Problem}
\label{sec:TEP}

The goal of table explicitation is to transform
an unstructured or semi-structured input string $x$, which is likely
lacking explicit row/column delimiters, into a string $y$ that is a structured representation
of a table. In our experiments, the output $y$ will be a string representing a table in HTML.
There are several possible sources for $x$, but
our immediate interest is in the text string $x$ that is stored in the clipboard when a user 
copies a table from any document, typically a PDF document. 
The clipboard data has no structural information of the table and this makes the
task of reliably retrieving the correct table quite challenging.
% formatting from copy-paste, flattened PDFs, emails, and reports, which causes traditional parsers to fail. 

For example, as shown in Figure~\ref{fig:overview}(b), a copied table text loses
all information about structure of the source table (in Figure~\ref{fig:overview}(a).
The goal of the table explicitation problem is to start with the text in Figure~\ref{fig:overview}(b) and regenerate a representation of the original table in Figure~\ref{fig:overview}(a).
%output T recovers the intended
%tabular structure in clean JSON (partial example:)

We solve the table explicitation problem using all three approaches: \\
(a) self-debug, \\
(b) end-to-end fine-tuning, and  \\
(c) last-mile fine-tuning.   \\
We compare these approaches and understand the trade-offs. 

Let us say we have some training data $(X_{train}, Y_{train})$ for the end-to-end table explicitation task.
We use a suitably-prompted pre-trained LLM to solve the table explicitation task, and let $f$ be the 
function represented by such an LLM. Then, the training data for the last-mile repair task consists
of $((X_{train}, f(X_{train})), Y_{train})$.
As our pre-trained LLM we use Gpt-4o. 
We perform our experiments using several different SLMs; namely,
(1) the 1B and 3B models from the Llama 3.2 family~\citep{grattafiori2024llama}, 
(2) the 4B model from the Qwen3 family~\citep{yang2025qwen3}, 
(3) version 3 of the 7B model and the 24B mistral small from Mistral AI~\citep{Mistral_AI_Team_2025, Jiang2023Mistral7}, 
and
(4) the base model from the phi-4 family~\citep{Abdin2024Phi3TR, abdin2024phi}. 
% and 
% (5) the mini model with 128k context length from the phi-3 family~\citep{Abdin2024Phi3TR, abdin2024phi}. 

\section{Dataset and Metrics}
\label{sec:datasets}
        We make use of three different datasets in our experiments, those being PubTabNet, FinTabNet and SciTSR. 

        \begin{enumerate}
            \item 
                PubTabNet~\citep{zhong2020imagebasedtablerecognitiondata}: A collection of Images and HTML-encoded tables, totaling to about 568,000 tables. The Images are extracted from the PubMed Central Open Access Subset (PMCOA), with the HTML representations coming from the XML representations of the tables present in the PMCOA. Each individual table was converted into a standalone HTML document, which was in turn converted to a PDF from which the raw clipboard text was extracted.
            \item 
                FinTabNet~\citep{lysak2023optimizedtabletokenizationtable}: FinTabNet OTSL is a collection of about 70k tables taken from annual reports of companies tracked by the S\&P 500, adapated from FinTabNet by creating a new markup language for table recognition. The HTML annotations served as the groundtruth, with the raw text extracted in the same manner as PubTabNet.
            \item 
                SciTSR~\citep{chi2019complicatedtablestructurerecognition}: A collection of 15000 PDFs and corresponding structural annotations in JSON. The JSON tables contain content as well and cell spanning information, which can procedurally be converted into an HTML table. Since the PDFs themselves are already standalone tables here, the raw text can be extracted directly from them. 
        \end{enumerate}
        We took a random sample of 15515 table explicitation tasks from the above three datasets. We tried to extract a table for each task using GPT-4o. We used a simple programmatic checker to determine if the extracted table was malformed, or a very poor quality table that was not close to the ground truth. If so, then we removed that task from our collection because it is not a candidate for last-mile repair. The remaining tasks were used to get a train-validation-test split.  Specifically, we obtained a train/val/test split of 8967/1133/2596 samples this way; see Table~\ref{tab:dataset_split} for contribution of each dataset to this final set. % Outputs that were not well-formed were removed from the dataset. 
        % Only used for User Interaction, omitted from paper.
        % Additonally, when exhaustive evaluation on the full dataset was not feasible due to high inference costs, we also created a smaller subset of the data with 90 samples using the technique described in~\citet{DBLP:journals/corr/abs-2503-10698}, using all-mpnet-base-v2 as the embedding model~\citep{reimers-gurevych-2019-sentence, NEURIPS2020_c3a690be}. 

        \ignore{ %%%%%%%%%%%%%%%%%%%%%%%%%% IGNORING FOR NOW %%%%%%%%%%%%%%%%%%%%%%%%%%
        \subsection{Repair by Supervised Finetuning from Zero-Shot Extraction}
            Let $r$ denote the raw text extracted from a PDF document, $c$ denote the coarse table extracted by the LLM and $g$ be the groundtruth formatted table. A small model is fine-tuned to map tuples of ($r$, $c$) $\mapsto$ $g$. The idea behind this is that small models are much more resource efficient finetune, deploy and maintain.
        
        \subsection{Repair by Learning in-context from User Inputs}
            Language models have shown the capacity to learn from examples in-context. Given that a user needs to make edits $e_1, e_2\dots e_n$ and table states $t_0, t_1, t_2 \dots t_n$ correspond to the different edits the user makes, the goal of this is to give a model the raw text, the latest table state as well as the sequence of edits leading up to it and predict a table as close to $t_n$ as possible. The workflow was exactly as follows:
            \begin{enumerate}
                \item 
                    The user makes an edit, which is added to the current queue of edits
                \item 
                    The model is called with the raw text, the current state and edit(s) made
                \item 
                    The prediction is checked to see if it preserves all of the user edits present in the queue
                \item 
                    If yes, the edited table is presented to the user. If the user rejects the prediction or any edit made by the user is not preserved, return to step 1. If the edited table is accepted, the queue is cleared.
            \end{enumerate}
            To simulate the behaviour of a user, a proxy program was created, which had access to the groundtruth table. At every step of the program, it would find the earliest mismatched cell (top-leftmost) and replace it using the operations provided by the TED computation against the groundtruth. In addition to checking if the edits in the queue were preserved, the proxy also checks if the model has made any additional progress beyond the user edits, because the user would not accept a predict table otherwise. This is primarily to guard against models predicting tables with no changes, which would empty the edit queue after marking the prediction as accepted.
            % To simulate the behaviour of a user, a proxy program was created, which had access to the groundtruth table. At every step of the program, it would find the earliest mismatched cell (top-leftmost) and replace it either a. using the operations provided by the TED computation against the groundtruth or b. to match the groundtruth exactly. In addition to checking if the edits in the queue were preserved, the proxy also checks if the model has made any additional progress beyond the user edits, because the user would not accept a predict table otherwise.
            \endignore}
        
        \subsection{Metrics}
        We need to compare the predicted table $\hat{y}$ with the ground-truth table $y$, where both tables are represented in HTML. A naive metric would be exact string equality, or {\em{exact match}} (EM), but there are more refined metrics that provide a more fine-grained view on the performance of an approach.

            \subsubsection{Tree Edit Distance (TED)}
                Since the ground-truth and the generated table are both HTML strings, they can both be represented as trees using the HTML tags as node labels. We can use tree edit distance, $\mathtt{Ted}(t_1,t_2)$, to compute the distance between two such trees, say $t_1$ and $t_2$, where $\text{\textsc{Ted}}(t_1,t_2)$ is the least number of replacements, deletions and additions needed to transform $t_1$ to $t_2$, and can be found using the Zhang-Shasha algorithm~\cite{TED}. This can be converted to a similarity score between 0 and 1, called {\em{Tree Edit Distance Similarity}}, or TEDS, using the formula 
                $\text{\textsc{Teds}}(t_1,t_2) = 1 - \mathtt{Ted}(t_1,t_2)/(|t_1|+|t_2|)$, 
                where $|t_i|$ denotes the size (number of nodes) of tree $t_i$. 
             Note that TEDS counts replacement of string $s$ by string $s'$ as 1 edit. 
             Since not all string edits require equal effort from the user, we can modify 
             the TED distance between two labels from being a 0-1 value to 
             to being the Levenshtein or String edit distance. This gives us the $\mathtt{Lev-Ted}$
             distance metric.
             While the $\mathtt{Lev-Ted}$ metric may seem more reflective of how much effort the user has to make
             (to transform tree $t_1$ to $t_2$), it is worth noting that it can be heavily skewed by outliers 
             (benchmarks with large content strings). Hence, we report both numbers.
             Higher is better for TEDS similarity and lower is better for the Lev-Ted metric since that would
             indicate that the predicted tree $t_1$ is closer to ground-truth tree $t_2$.
             % of it being \textcolor{red}{unbounded in the positive direction}, and the definition of TED does not allow for copying content from one cell into another, so neither metric can be looked at in isolation.
            \subsubsection{Grid Table Similarity (GriTS)}
                GriTS is a metric introduced by~\citet{smock2023gritsgridtablesimilarity}. This processes the matrix representations of both tables and finds a lower and upper bound on the 2D most similar substructure between both matrices, which is a relaxation of the 2D most common substructure problem. Solving this problem using different similarity functions for topology and content leads to two different metrics, GriTS$_{top}$ and GriTS$_{con}$ measuring the similarity of table topology and content respetively. These metrics are computed as a pair of values, a lower bound and an upper bound, and the true value is in between those values. Since the two bounds were very close in our experiments, we report only the mean of the bounds. Since these are similarity measures, higher is better for these metrics.
                
\section{Evaluation}
\label{sec:eval}

We now evaluate the \lift approach by focusing on three key questions:
\\
(1)  How does \lift compare to self-debug and \etoe approaches on the table explicitation task?
We clearly understand the trade-off in terms of the compute cost -- because we know exactly how many
LLM calls and how many SLM calls each technique uses -- so, we focus here purely on how well each approach
does on the metrics outlined above.
\\
(2) How does the limited availability of training data impact \lift and \etoe approaches?
\\
(3) Are \lift and \etoe both equally robust to changes in the input text?

\begin{table}[t]
\begin{center}
\begin{tabular}{l|rrr|rrr}
\toprule
    & \multicolumn{3}{c|}{\bf TEDS} 
    & \multicolumn{3}{c}{\bf LevTED} \\
    \multicolumn{1}{c|}{\bf Model}  
    &\multicolumn{1}{c}{$\text{\sd}(M)$} 
    &\multicolumn{1}{c}{$\text{\lift}(M)$} 
    &\multicolumn{1}{c|}{$\text{\etoe}(M)$}
    &\multicolumn{1}{c}{$\text{\sd}(M)$} 
    &\multicolumn{1}{c}{$\text{\lift}(M)$} 
    &\multicolumn{1}{c}{$\text{\etoe}(M)$} \\
\midrule
    Llama 3.2 1B & 0.696 & 0.880 & 0.887 & 477 & 214 & 192 \\ 
    Llama 3.2 3B & 0.743 & 0.903 & 0.905 & 371 & 148 & 133 \\
    Qwen 3 4B & 0.835 & 0.901 & 0.838 & 227 & 136 & 132 \\
    Mistral 7B & 0.761 & 0.936 & 0.927 & 377 & 98 & 112 \\
    Phi-4 14B & 0.855 & 0.923 & 0.834 & 186 & 104 & 152 \\
    Mistral 24B & 0.946 & 0.951 & 0.945 & 74 & 66 & 71 \\
    Gpt-4o & 0.873 & -- & -- & 144 & -- & -- \\
\bottomrule
\end{tabular}
\end{center}
    \caption{{\small{Comparing the three approaches, self-debug (\sd), last-mile fine-tuning (\lift) and end-to-end fine-tuning (\etoe) on table explicitation tasks across different choices for underlying repair SLM using the TEDS similarity (higher is better) and LevTed distance (lower is better) metrics.}}}\label{tab:versus}
\end{table}

\subsection{Comparing the three approaches}
\label{sec:rq1}
We fine-tuned six SLMs, namely 
Llama 3.2 1B, 
Llama 3.2 3B,
Qwen 3 4B,
Mistral 7B,
Phi-4 14B, and
Mistral 24B, on two tasks separately: (a) the last-mile repair task and (b) end-to-end table explicitation task.
The input for last-mile repair task was generated by using Gpt-4o on the input of the table explicitation task, as described in
Section~\ref{sec:TEP}.  The models were all fine-tuned for 5 epochs, with an effective batch size of 128. A total of 50 warmup steps up to a learning rate of $5\times 10^{-5}$ were used. The models were evaluated on the validation set after every epoch, and at the end of the training run the checkpoint with the smallest validation loss was chosen. Training was done at half-precision using the bfloat16 format and an 8-bit version of the Adam optimizer. LORA was used with rank 16 and $\alpha = 32$. The prompts we used can be found in Appendix~\ref{app:prompts}.

We then compared three approaches:\\
(1) self-debug, $\text{\sd}(M)$, where the base (not fine-tuned) model $M$ is used to perform the repair
on the table extracted by Gpt-4o, \\
(2) last-mile fine-tuning, $\text{\lift}(M)$, where the version of $M$ fine-tuned on last-mile repair is used to perform the repair on the table extracted by Gpt-4o, and \\
(3) end-to-end fine-tuning, $\text{\etoe}(M)$, where $M$ that is fine-tuned on the end-to-end table explicitation task is used to extract the table from the input text (no Gpt-4o is used in this scenario).\\

Table~\ref{tab:versus} shows data comparing the three approaches on table explicitation tasks across different choices for underlying repair SLM using the TEDS similarity (higher is better) and LevTed distance (lower is better) metrics.
We did not perform last-mile fine-tuning or the end-to-end fine-tuning on Gpt-4o and hence the blanks 
in Table~\ref{tab:versus}.
We observe that 
\\
(a) Across both metrics and across all models $M$, $\text{\lift}(M)$ is superior to $\text{\sd}(M)$, which should be expected since we are using a fine-tuned version of $M$ to repair in $\text{\lift}(M)$, whereas we use the base version of $M$ in 
$\text{\sd}(M)$.
\\
(b) Last-mile fine-tuning is a promising alternative to doing full task-specific fine-tuning. Both
$\text{\lift}(M)$ and $\text{\etoe}(M)$ perform comparably, and on the TEDS metric, in fact, $\lift(M)$ is either 
very close or arguably better than $\etoe$ for all models. Moreover, last-mile fine-tuning has other benefits, as we
will discuss later.
\\
(c) In general, as the SLMs get bigger, their performance {\em{generally}} improves across all approaches -- with
only Phi-4 breaking the trend (on both metrics) for $\text{\lift}(M)$ and $\text{\etoe}(M)$, and
only Mistral 7B breaking the trend (on both metrics) for $\text{\sd}(M)$. 
We also note
that \sd with Mistral 24B outperforms
\sd with Gpt-4o, which is surprising, but digging deeper into it, we observed that Mistral 24B
was frequently generating empty outputs, and we ignored those tasks when computing the metrics.
Table~\ref{tab:results_finetune} provides the raw count of how many tasks produced valid tables where
we see that \sd with Mistrat 24B produced 1625 non-empty and valid tables compared to around 2500 for
all other models. 
%perhaps points to the fact that a {\em{different}} model, even if it is
%less powerful, may do better at repair than the same model (that made the error in the first place). However, this
%only happens for Mistral 24B, so there is some minimum threshold required for the untrained small repair model to outperform
%the more powerful model (Gpt-4o) that performs the task originally.

The GriTS metrics were correlated with the TEDS metric and hence we did not show those numbers in Table~\ref{tab:versus}.
The results for GriTS metrics can be found in Table~\ref{tab:versus2} in the appendix, along with the 
consolidated full results in Table~\ref{tab:results_finetune}.

\subsection{Limited Training Data}

We next consider the scenario when there is limited training data. The hypothesis is that when training
data is scarce, the \lift approach would outperform the \etoe approach since the last-mile repair
task is presumably simpler than the end-to-end table explicitation task. To test this hypothesis,
we pick two base models, Mistral-7B and Llama 3.2 3B, and fine-tune them each with the same number of
training data points -- we fine-tune each model once for the last-mile repair task (and use that model
in \lift) and once for the data explicitation task (and use that model for \etoe evaluation).

\begin{table}[t]
\begin{center}
\begin{tabular}{l|l|rr||l|rr}
\toprule
    Train & Base
    & \multicolumn{2}{c|}{\bf{TEDS}}
    & Base
    & \multicolumn{2}{c}{\bf{TEDS}} \\
    Set Size & Model
    &\multicolumn{1}{c}{$\text{\lift}(M)$} 
    &\multicolumn{1}{c|}{$\text{\etoe}(M)$}
    & Model
    &\multicolumn{1}{c}{$\text{\lift}(M)$} 
    &\multicolumn{1}{c}{$\text{\etoe}(M)$} \\
\midrule
    1000 & Mistral-7B & 0.875 & 0.731 & Llama3.2-3B & 0.710 & 0.740 \\
    2000 & Mistral-7B & 0.878 & 0.782 & Llama3.2-3B & 0.823 & 0.784 \\
    4000 & Mistral-7B & 0.916 & 0.832 & Llama3.2-3B & 0.879 & 0.841 \\
    6000 & Mistral-7B & 0.920 & 0.858 & Llama3.2-3B & 0.891 & 0.839 \\
\bottomrule
\end{tabular}
\end{center}
    \caption{{\small{Comparing the last-mile fine-tuning (\lift) and end-to-end fine-tuning (\etoe) for a fixed size of training set (higher TEDS score is better).}}}\label{tab:trainset}
\end{table}

Table~\ref{tab:trainset} shows the results. We show only the TEDS metric since the other metrics showed the
same general trends. The full results are included in the Appendix in Table~\ref{tab:size_ablation}.
When using the same number of training data points, the \lift approach mostly outperforms 
\etoe, which validates our hypothesis.  For Mistral-7B we see a nice trend of the performance
gap shrinking as the training set size grows from 1000 to 6000. That trend, however, does
not hold for Llama, where there is no clear pattern in the performance gap. In fact, surprisingly, 
with 1000 training points, \etoe performs slightly better when Llama is the base model.
In general, though, the results show that it is beneficial to use \lift when there is limited
training data (and when using a large language model at the first step is an option).

We also note here that there is a cost (and latency) tradeoff between the three approaches
\sd, \lift and \etoe.
The \sd approach has no additional training cost, but requires one LLM call for table explicitation,
and another LLM or SLM call for repair at test time.
Both \lift and \etoe require some amount of training, though \lift 
requires less of it as we saw above. At test time, \lift requires one LLM call and one SLM call
whereas \etoe requires only one SLM call. So, purely from cost and latency point of view, \etoe
is the best, but \lift provides certain other benefits.

\subsection{Robustness to Input Format Variability}
\label{sec:noisy}

A benefit of using a general-purpose LLM is that it can handle inputs that are slightly different
from those in the training data without a significant drop in its accuracy.
Since the \lift approach uses an LLM in its first step, a natural question is whether \lift inherits some
of this robustness to input format variability, and how does it compare with \etoe on this aspect.

To answer the above question, we created inputs in two different ``formats''. 
We randomly sampled 100 datapoints from the test set.
We converted the ground-truth tables into broken \textsc{Csv} and well-formed \textsc{Json} files.  
Ground-truth tables with spanning cells were excluded because there is no canonical way of representing 
spans (adjacent cells merged into one big cell) in either format. 
Since \textsc{Json} representation requires column headers, any ground-truth table that did not contain a header 
($\langle$thead$\rangle$) was excluded when creating noisy \textsc{Json}s. 
The noisy \textsc{Csv} were generated by starting with a clean \textsc{Csv} and 
(1) randomly changing some separator to a comma, tab, pipe or caret, and
% The CSVs, at random had their separators changed between commas, tabs, pipes and carets, and 
(2) adding data to the header or footer of a \textsc{Csv}. 
The \textsc{Json}s were just well-formed \textsc{Json} representations of ground-truth tables. Such \textsc{Json}s can be
easily converted to \textsc{Html} tables by a few lines of code, but the noisy \textsc{Csv}s are not so easily translated to 
\textsc{Html} tables by code. However, for our experiments, they provide two good tests for robustness to
format variability because they both are different from our train and test sets for 
table explicitation.

\begin{table}[t]
\begin{center}
\begin{tabular}{l|rr|rr}
\toprule
    & \multicolumn{2}{c|}{\bf{CSV}}
    & \multicolumn{2}{c}{\bf{JSON}} \\
    Model M 
    &\multicolumn{1}{c}{$\text{\lift}(M)$} 
    &\multicolumn{1}{c|}{$\text{\etoe}(M)$}
    &\multicolumn{1}{c}{$\text{\lift}(M)$} 
    &\multicolumn{1}{c}{$\text{\etoe}(M)$} \\
\midrule
    Llama 3.2 1B & 0.844 & 0.643 & 0.785 & 0.504 \\ 
    Llama 3.2 3B & 0.774 & 0.769 & 0.503 & 0.508 \\
    Qwen 3 4B & 0.869 & 0.751 & 0.630 & 0.695 \\
    Mistral 7B & 0.762 & 0.751 & 0.673 & 0.751 \\
    Phi-4 14B & 0.904 & 0.831 & 0.750 & 0.652 \\
    Mistral 24B & 0.844 & 0.810 & 0.820 & 0.809 \\
\bottomrule
\end{tabular}
\end{center}
    \caption{{\small{Comparing the last-mile fine-tuning (\lift) and end-to-end fine-tuning (\etoe) on differently-formatted inputs - broken CSV and well-formed JSON using TEDS (higher score is better).}}}\label{tab:noisy}
\end{table}

Table~\ref{tab:noisy} contains the results of evaluating \lift and \etoe on noisy \textsc{Csv} and clean \textsc{Json} inputs. 
We used the same model checkpoints that were used in earlier experiments reported in Section~\ref{sec:rq1}.
Compared to how they performed in Table~\ref{tab:versus}, 
both approaches performed worse here. On the one hand, this is expected since the inputs
here are ``out-of-distribution'' compared to what the models were trained on.
On the other hand, even though the inputs are different here, they actually have more information
than the text clipboard dumps from PDF selections since there is some structural information in these
inputs, but that does not seem to be helping either approach as much. 

The main take-away from Table~\ref{tab:noisy} is that \lift performs consistently better than \etoe on the 
broken \textsc{Csv}s across the whole set of SLMs.
This serves as evidence for the hypothesis that last-mile fine-tuning is more robust than 
end-to-end fine-tuning.
If we look at the performance on \textsc{Json}, there is no clear winner across models. 
This is probably because \textsc{Json} is a standard format and 
we didn't inject any noise or errors in the \textsc{Json}s. Different SLMs may have different expertise with manipulating
\textsc{Json} format, which may be reflected in the results of Table~\ref{tab:noisy}.

\ignore{ %%%% ASHISH: COMMENTING THIS OUT
        \subsection{Learning from User Examples}
            For this experiment, we were attempting to fix the tables generated by Mistral 7B. The user proxy was allowed to make 5 edits, with the models making a prediction after each round. The repairs were run over 20 tables, with any tables fixed by the proxy before the model makes a prediction being discarded. The results are in Table \ref{tab:repair_results}
            % \begin{table}[ht]
            %     \begin{adjustbox}{center}
            %         \begin{tabular}{|c|c|c|c|c|}
            %             \hline
            %             Model & Acceptance Ratio & Tables Fixed & Rounds To Fix & Roundwise Lev-TED Drop \\
            %             \hline
            %             GPT 4o & 0.462 & 6 & 5 & 50.210 \\
            %             \hline
            %             Mistral 7B (no sft) & 0.159 & 3 & 2.33 & 75.18 \\
            %             \hline
            %             Mistral 7B (repair sft) & 0.110 & 4 & 1.75 & 100.28 \\
            %             \hline
            %         \end{tabular}        
            %     \end{adjustbox}
            %     \caption{Results for Table Repair from User Edit Examples}
            %     \label{tab:repair_results}
            % \end{table}

            \begin{table}[ht]
                \centering
                \begin{tabular}{|c|c|c|c|c|}
                     \hline
                     Model & Unmodified Prediction & Edit Not Preserved & Acceptances & Average TEDS improvement\\
                     \hline
                     Gpt-4o & 0.121 & 0.179 & 0.177 & \\
                     \hline
                     LLama 3.2 1B & 0.000 & 0.990 & 0.005 \\
                     \hline
                     LLama 3.2 3B & 0.019 & 0.883 & 0.027 \\
                     \hline
                     Mistral-7B & 0.049 & 0.521 & 0.142 \\
                     \hline
                     Qwen3 - 4B & 0.213 & 0.500 & 0.154 \\
                     \hline
                     Phi-4 &  0.075 & 0.378 & 0.188 \\
                     \hline
                \end{tabular}
                \caption{Results for Table Repair from User Edit Examples}
                \label{tab:repair_results}
            \end{table}

            On aggregate, the LLM seems to perform the best, with the lowest rate of predictions that undo the user edits, a common problem with neural in-line completion solution. It also has the second lowest acceptance rate, but by a very slim margin. This is also aided by the fact that the model outperforming it has more than twice the rate of predictions undoing user edits. 
            \endignore}
%%%%%%%%%%%%%%%%%%%%%%%%%%%%%%%%%%%%%%%%%%%%%%%%%%%%%%% END OF COMMENT

    \section{Conclusion}
        In this paper, we explored the table explicitation task and experimentally evaluated three possible approaches for it;
        namely, 
        (a) self-debug, where a pre-trained LLM is used to solve the task and then (the same LLM or a different SLM) 
        is used to repair the output, 
        (b) last-mile fine-tuning, where an LLM is used to solve the task and then a fine-tuned
        SLM is used to repair the output, and
        (c) end-to-end fine-tuning, where an SLM is fine-tuned for the entire task. 
        A few clear trade-offs emerged.
        % each applicable to a different scenario. 
        When a large amount of training data is available, it is beneficial to fine-tune an SLM for the end-to-end task
        because this is the most cost-effective solution as it involves no LLM calls. 
        % fine-tuning and deploying a model is feasible, finetuning an SLM for table extraction is the optimal solution. 
        If data is limited and a certain amount of robustness to input variability is desired, the
        last-mile fine-tuning approach is most effective -- it does no worse than or better than end-to-end fine-tuning.  It also does better than self-debug and
        is also more cost-effective than self-debug with LLMs.
        % two step pipeline with an LLM and an LLM finetuned for repair is the best way to go. 
        If the option to fine-tune and deploy fine-tuned SLMs is not available, then using an LLM to solve the task and then using the same LLM or a powerful SLM
        to repair (without any fine-tuning) is a good option.

\bibliography{references}
\bibliographystyle{colm2026_conference}
    
\appendix
\section{Detailed Experimental Results}

\subsection{Comparing the three approaches}
Table~\ref{tab:results_finetune} shows the full results for the experiment comparing self-debug, 
last-mile file-tuning, and end-to-end fine-tuning. It includes the following baselines:
            \begin{enumerate}
                \item
                    Using GPT 4o to extract the table from the raw text
                \item 
                    Using GPT 4o to extract the table, and a second call to GPT 4o to repair the table, which we have called $\sd(Gpt4o)$.
            \end{enumerate}
            The results are shown in Table \ref{tab:results_finetune}, using the 1B and 3B models from the Llama 3.2 family \citep{grattafiori2024llama}, the 4B model from the Qwen3 family \citep{yang2025qwen3}, version 3 of the 7B model and the 24B mistral small from Mistral AI \citep{Mistral_AI_Team_2025, Jiang2023Mistral7}, 
            and
            the base model from the phi-4 family~\citep{Abdin2024Phi3TR, abdin2024phi}. 
            %and the mini model with 128k context length from the phi-3 family \citep{Abdin2024Phi3TR, abdin2024phi}. 
            In this table, ``\sd'' denotes self-debug where the table explicitation
            task is first solved by Gpt-4o, and its result is repaired using either 
            Gpt-4o (Row 2) or an  SLM {\em{with no fine-tuning}}.
            The term ``\lift'' denotes last-mile fine-tuning, where Gpt-4o again solves the table explicitation task first
            and then its output is repaired using a fine-tuned SLM.
            The term ``\etoe'' denotes the end-to-end fine-tuning approach, where the end-to-end task is performed using a 
            fine-tuned model.
            We only fine-tuned SLMs, so there are no results for \lift(Gpt-4o) or \etoe(Gpt-4o).

            % We additionally attach the results for experiments where the small models were asked to the end-to-end task rather than last-mile repair, marked as (extraction).

            % The baseline had no spans, so all the models had to insert any and all spans. 
            \begin{table}[ht]
                \begin{adjustbox}{center}
                    \begin{tabular}{|c|c|c|c|c|c|}
                        \hline
                        Model & $\frac{\mbox{Exact Match}}{\mbox{Valid Tables}}$ & GriTS$_{top}$ & GriTS$_{con}$ & TEDS & Lev-TED \\
                        \hline
                        GPT 4o (extraction)  & 166/2596 & 0.678 - 0.678 & 0.561 - 0.569 & 0.686 & 491.435 \\
                        \hline 
                        GPT 4o (\sd) & 710/2571 & 0.899 - 0.899 & 0.842 - 0.846 & 0.873 & 143.877 \\
                        \hline \hline

                        Llama 3.2 1B (\sd) & 66/2485 & 0.679 - 0.679 & 0.539 - 0.548 & 0.696 & 477.304 \\
                        \hline
                        Llama 3.2 3B (\sd) & 182/2425 & 0.732 - 0.732 & 0.602 - 0.610 & 0.743 & 370.878 \\
                        %\hline
                        %Phi-3-mini (\sd) & 0/152 & 0.255 - 0.255 & 0.096 - 0.106 & 0.282 & 2069.290 \\
                        \hline
                        Qwen 3 4B (\sd) & 424/2470 & 0.825 - 0.825 & 0.747 - 0.752 & 0.835  & 226.900 \\
                        \hline
                        Mistral-7B (\sd) & 262/2494 & 0.760 - 0.761 & 0.629 - 0.638 & 0.761 & 377.253 \\
                        \hline
                        Phi-4 (\sd) & 500/2586 & 0.862 - 0.862 & 0.787 - 0.792 & 0.855  & 186.273 \\
                        \hline
                        Mistral-24B (\sd) & 319/1625 & 0.852 - 0.852 & 0.771 - 0.776 & 0.881 & 184.952 \\
                        \hline \hline

                        Llama 3.2 1B (\lift) & 552/2581 & 0.841 - 0.841 & 0.745 - 0.752 & 0.880 & 214.420 \\
                        \hline
                        Llama 3.2 3B (\lift) & 782/2588 & 0.878 - 0.878 & 0.811 - 0.816 & 0.903 & 148.014 \\
                        %\hline
                        %Phi-3-mini (\lift) & 0/382 & 0.232 - 0.233 & 0.086 - 0.096 & 0.346 & 1528.552 \\
                        \hline
                        Qwen 3 4B (\lift) & 696/2585 & 0.882 - 0.883 & 0.816 - 0.821  & 0.901 & 136.280 \\
                        \hline
                        Mistral-7B (\lift) & 1232/2577 & 0.926 - 0.926 & 0.879 - 0.881 & 0.936 & 97.610 \\
                        \hline
                        Phi-4 (\lift) & 1006/2596 & 0.915 - 0.915 & 0.870 - 0.873  & 0.923 & 103.786 \\
                        \hline
                        Mistral-24B (\lift) & 1488/2579 & 0.951-0.951 & 0.924 - 0.925 & 0.951 & 65.580 \\
                        \hline\hline

                        Llama 3.2 1B (\etoe) & 552/2568 & 0.855 - 0.855 & 0.763 - 0.769 & 0.887 & 192.132 \\
                        \hline
                        Llama 3.2 3B (\etoe) & 812/2587 & 0.888 - 0.888 & 0.821 - 0.826 & 0.905 & 132.714 \\
                        %\hline
                        %Phi-3-mini (\etoe) & 768/2389 & 0.883 - 0.883 & 0.816 - 0.820 & 0.906 & 140.931 \\
                        \hline
                        Qwen 3 4B (\etoe) &  729/2586 & 0.884 - 0.885 & 0.820 - 0.825 & 0.838 & 131.920 \\
                        \hline
                        Mistral-7B (\etoe) & 1408/2591 & 0.917-0.917 & 0.866-0.867 & 0.927 & 112.272 \\
                        \hline
                        Phi-4 (\etoe) &  912/2594 & 0.887 - 0.887 & 0.832 - 0.832 & 0.834 & 152.23 \\
                        \hline
                        Mistral-24B (\etoe) & 1408/2591 & 0.942-0.943 & 0.911-0.912 & 0.945 & 70.71 \\
                        \hline
                    \end{tabular}
                 \end{adjustbox}
                \caption{{\small{Consolidated results for table explicitation problem using last-mile repair (labeled ``\lift''), self-debug (labeled ``\sd''), and end-to-end fine-tuning (labeled ``\etoe''). The first row is the Gpt-4o baseline (single LLM call). Not all methods successfully completed all tests and the denominator in 2nd column shows that. Upper and lower bounds for GriTS scores are provided. All models are instruction tuned.}}}
                \label{tab:results_finetune}
            \end{table}

            The small models see major improvements in extracting both the topology and the content from supervised finetuning. %Additionally, the fact that the out of the box SLMs both outperform GPT 4o in repairs show that the table repair task is in fact easier than a generation-from-scratch task. 
            Mistral 24B appears to have some interesting statistics because even without fine-tuning, its TEDS
            score 0.946 (for Gpt-4o extraction + Mistral 24B base repair) is better than the TEDS score
            0.876 that Gpt-4o gets (when it does both extraction and repair step). 
            In fact, Mistral 24B base outperforms even the fine-tuned version of it's 7B counterpart in terms of Levenshtein TED.
            However, we also note that
            Mistral 24B base (without fine-tuning) generates valid HTML much less frequently 
            (1625) than any other model tested (all others have around 2500) and this could have contributed to the
            skew in numbers. 
            %This is possibly due to the model making indicates the Mistral'
            %This is possibly happening
            %because of gains due to ``diversity'' coming from the repair model being different  from the extractor model.
            % The only other outlier is phi-3-mini, which even performs exceptionally poorly after supervised fine-tuning. 
            However, the general trends are within expectations, with a loose correlation between model size and performance, and SFT improves the number of valid HTML tables generated, as well as improving the performance on all of the measured metrics.
            
We had extracted Table~\ref{tab:versus} from Table~\ref{tab:results_finetune}.
Table~\ref{tab:versus2} is identical to Table~\ref{tab:versus} except that it reports the GriTS metrics,
whereas Table~\ref{tab:versus} reported the TEDS and LevTed metrics.

\begin{table}[t]
\begin{center}
\begin{tabular}{l|rrr|rrr}
\toprule
    & \multicolumn{3}{c|}{$\mathbf{GriTS}_{top}$}
    & \multicolumn{3}{c}{$\mathbf{GriTS}_{con}$}  \\
    \multicolumn{1}{c|}{\bf Model}  
    &\multicolumn{1}{c}{$\text{\sd}(M)$} 
    &\multicolumn{1}{c}{$\text{\lift}(M)$} 
    &\multicolumn{1}{c|}{$\text{\etoe}(M)$}
    &\multicolumn{1}{c}{$\text{\sd}(M)$} 
    &\multicolumn{1}{c}{$\text{\lift}(M)$} 
    &\multicolumn{1}{c}{$\text{\etoe}(M)$} \\
\midrule
    Llama 3.2 1B & 0.679 & 0.841 & 0.855 & 0.544 & 0.749 & 0.766 \\ 
    Llama 3.2 3B & 0.732 & 0.878 & 0.888 & 0.606 & 0.814 & 0.824 \\
    Qwen 3 4B & 0.825 & 0.882 & 0.884 & 0.750 & 0.819 & 0.823 \\
    Mistral 7B & 0.760 & 0.926 & 0.917 & 0.634 & 0.880 & 0.867 \\
    Phi-4 14B & 0.862 & 0.915 & 0.887 & 0.790 & 0.872 & 0.832 \\
    Mistral 24B & 0.852 & 0.951 & 0.942 & 0.774 & 0.925 & 0.912 \\
\bottomrule
\end{tabular}
\end{center}
    \caption{{\small{Comparing the three approaches, self-debug (\sd), last-mile fine-tuning (\lift) and end-to-end fine-tuning (\etoe) on table explicitation tasks across different choices for underlying repair SLM using the GriTS topology and GriTS content metrics (higher is better).}}}\label{tab:versus2}
\end{table}

\subsection{Limited Training Data}
            We find that when data is limited, last-mile fine-tuning-based repair outperforms 
            end-to-end fine-tuning-based generation. 
            To this end, we obtain different model checkpoints for two models, Llama 3.2 3B and Mistral 7B. We train these SLMs  on both tasks (generation and repair) with different training set sizes (1000, 2000, 4000, 6000). The full results are shown 
            in~Table~\ref{tab:size_ablation}. The repair models, at every size, outperform their generation counterparts on almost every single metric at any amount of data. Llama 3.2 instruct at 1000 is the only exception, doing better at repair than generation in terms of TEDS and both GriTS metrics. 
             \begin{table}[ht]
                \begin{adjustbox}{center}
                    \begin{tabular}{|c|c|c|c|c|c|}
                        \hline
                        Mistral-7B-Instruct-v0.3 & GriTS$_{top}$ & GriTS$_{con}$ & TEDS & Lev-TED \\
                        \hline
                        \etoe, 1000 datapoints  & 0.699 - 0.699 & 0.558 - 0.566 & 0.731 & 409.183 \\
                        \hline
                        \etoe, 2000 datapoints  & 0.745 - 0.745 & 0.636 - 0.642 & 0.782 & 330.438 \\
                        \hline
                        \etoe, 4000 datapoints  & 0.791 - 0.791 & 0.713 - 0.718 & 0.832 & 253.908 \\
                        \hline
                        \etoe, 6000 datapoints  & 0.826 - 0.826 & 0.744 - 0.749 & 0.858 & 222.347 \\
                        \hline
                        \lift, 1000 datapoints  & 0.832 - 0.832 & 0.730 - 0.739 & 0.875 & 211.231 \\
                        \hline
                        \lift, 2000 datapoints  & 0.833 - 0.833 & 0.741 - 0.749 & 0.878 & 196.384 \\
                        \hline
                        \lift, 4000 datapoints  & 0.899 - 0.899 & 0.836 - 0.841 & 0.916 & 128.148 \\
                        \hline
                        \lift, 6000 datapoints  & 0.909 - 0.909 & 0.856 - 0.859 & 0.920 & 122.961 \\
                        \hline\hline
                        Llama-3.2-3B-Instruct & GriTS$_{top}$ & GriTS$_{con}$ & TEDS & Lev-TED \\
                        \hline
                        \etoe, 1000 datapoints  & 0.688 - 0.688 & 0.559 - 0.567 & 0.740 & 439.855 \\
                        \hline
                        \etoe, 2000 datapoints  & 0.722 - 0.723 & 0.609 - 0.617 & 0.784 & 379.255 \\
                        \hline
                        \etoe, 4000 datapoints  & 0.789 - 0.789 & 0.707 - 0.713 & 0.841 & 270.541 \\
                        \hline
                        \etoe, 6000 datapoints  & 0.790 - 0.791 & 0.710 - 0.717 & 0.839 & 283.733 \\
                        \hline
                        \lift, 1000 datapoints  & 0.650 - 0.650 & 0.517 - 0.528 & 0.710 & 410.683 \\
                        \hline
                        \lift, 2000 datapoints  & 0.759 - 0.760 & 0.630 - 0.639 & 0.823 & 274.262 \\
                        \hline
                        \lift, 4000 datapoints  & 0.838 - 0.839 & 0.741 - 0.749 & 0.879 & 194.125 \\
                        \hline
                        \lift, 6000 datapoints  & 0.863 - 0.864 & 0.778 - 0.785 & 0.891 & 167.004 \\
                        \hline
                    \end{tabular}
                \end{adjustbox}
                 \caption{{\small{Comparison of \lift and \etoe approaches at different training dataset sizes. The \lift approach geenrally performs better than the \etoe approach acroos different train set sizes, with the gap being larger the lesser data there is.}}}
                \label{tab:size_ablation}
            \end{table}
            
\subsection{Robustness to Input Format Variability}
 Table~\ref{tab:results_gen_csv} shows the full results (all four metrics) for both \lift and \etoe approaches
 on broken CSVs, which was discussed in Section~\ref{sec:noisy}. 
 % In this table, ``generate'' refers to the \etoe approach and ``repair'' refers to the \lift approach.
 Table~\ref{tab:results_gen_json} shows the same data but for well-formed JSON inputs. Part of the results
 from these two tables was shown before in Table~\ref{tab:noisy}.
            \begin{table}[ht]
                \begin{adjustbox}{center}
                    \begin{tabular}{|c|c|c|c|c|c|}
                        \hline
                        Model & GriTS$_{top}$ & GriTS$_{con}$ & TEDS & Lev-TED \\
                        \hline
                        LLM & 0.966 - 0.966 & 0.923 - 0.924 & 0.925 & 53.76 \\
                        \hline
                        Mistral-7B (\etoe) & 0.822 - 0.822 & 0.663 - 0.667 & 0.751 & 248.386 \\
                        \hline
                        Mistral-24B (\etoe)& 0.883 - 0.884 & 0.802 - 0.803 & 0.810 & 165.75 \\
                        \hline
                        Llama 3.2 1B (\etoe)& 0.641 - 0.641 & 0.414 - 0.420 & 0.643 & 466.286 \\
                        \hline
                        Llama 3.2 3B (\etoe)& 0.824 - 0.885 & 0.672 - 0.676 & 0.769 & 262.220 \\
                        \hline
                        Qwen 3 4B (\etoe)& 0.879 - 0.879 & 0.790 - 0.793 & 0.751 & 160.46 \\
                        \hline
                        Phi-4 (\etoe)& 0.904 - 0.904 & 0.804 - 0.805 & 0.831 & 143.231 \\
                        % \hline
                        % Phi-3 mini (\etoe)& 0.770 - 0.770 & 0.635 - 0.637 & 0.556 & 289.298 \\
                        \hline
                        Phi-4 (\lift) & 0.945 - 0.945 & 0.882 - 0.883 & 0.904 & 88.505 \\
                        \hline
                        Qwen 3 4B (\lift) & 0.876 - 0.876 & 0.798 - 0.802 & 0.869 & 121.260 \\
                        \hline
                        Llama 3.2 1B (\lift) & 0.878 - 0.881 & 0.767 - 0.776 & 0.844 & 192.950 \\
                        \hline
                        Llama 3.2 3B (\lift) & 0.822 - 0.822 & 0.677 - 0.681 & 0.774 & 216.460 \\
                        %\hline
                        %Phi 3 mini (\lift) & 0.215 - 0.215 & 0.072 - 0.080 & 0.355 & 1105.000 \\
                        \hline
                        Mistral 7B (\lift) & 0.821 - 0.821 & 0.695 - 0.697 & 0.762 & 226.277 \\
                        \hline
                        Mistral 24B (\lift) & 0.934 - 0.934 & 0.843 - 0.844 & 0.844 & 108.283 \\
                        \hline
                    \end{tabular}
                 \end{adjustbox}
                \caption{{\small{Comparing \lift and \etoe on extracting tables from broken CSV files. Note that all models were still fine-tuned on the original training data coming from the clipboard when copying from PDF documents.}}}
                \label{tab:results_gen_csv}
            \end{table}

            \begin{table}[ht]
                \begin{adjustbox}{center}
                    \begin{tabular}{|c|c|c|c|c|c|}
                        \hline
                        Model & GriTS$_{top}$ & GriTS$_{con}$ & TEDS & Lev-TED \\
                        \hline
                        LLM & 0.880 - 0.880 & 0.783 - 0.784 & 0.826 & 256.142 \\
                        \hline
                        Mistral-7B (\etoe) & 0.503 - 0.503 & 0.243 - 0.263 & 0.751 & 793.021 \\
                        \hline
                        Mistral-24B (\etoe) & 0.803 - 0.803 & 0.662 - 0.674 & 0.809 & 293.244 \\
                        \hline
                        Llama 3.2 1B (\etoe) & 0.407 - 0.407 & 0.123 - 0.143 & 0.504 & 1027.071 \\
                        \hline
                        Llama 3.2 3B (\etoe) & 0.411 - 0.412 & 0.152 - 0.172 & 0.508 & 1044.437 \\
                        \hline
                        Qwen 3 4B (\etoe) & 0.617 - 0.617 & 0.443 - 0.460 & 0.695 & 499.795 \\
                        \hline
                        Phi-4 (\etoe) & 0.590 - 0.590 & 0.370 - 0.394 & 0.652 & 599.279 \\
                        %\hline
                        %Phi-3 mini (\etoe) & 0.452 - 0.452 & 0.175 - 0.193 & 0.498 & 775.095 \\
                        \hline
                        Phi-4 (\lift) & 0.708 - 0.708 & 0.573 - 0.587 & 0.750 & 413.111 \\
                        \hline
                        Qwen 3 4B (\lift) & 0.511 - 0.511 & 0.330 - 0.348 & 0.630 & 630.068 \\
                        \hline
                        Llama 3.2 1B (\lift) & 0.746 - 0.747 & 0.629 - 0.639 & 0.785 & 361.231 \\
                        \hline
                        Llama 3.2 3B (\lift) & 0.459 - 0.460 & 0.222 - 0.236 & 0.503 & 976.220 \\
                        %\hline
                        %Phi 3 mini (\lift) & 0.265 - 0.266 & 0.077 - 0.087 & 0.382 & 1620.727 \\
                        \hline
                        Mistral 7B (\lift) & 0.679 - 0.679 & 0.455 - 0.467 & 0.673 & 606.344 \\
                        \hline
                        Mistral 24B (\lift) & 0.811 - 0.813 & 0.686 - 0.699 & 0.820 & 226.495 \\
                        \hline
                    \end{tabular}
                 \end{adjustbox}
                %\caption{Results for extraction of JSON tables}
                \caption{{\small{Comparing \lift and \etoe on extracting tables from well-formed JSON files. Note that all models were still fine-tuned on the original training data coming from the clipboard when copying from PDF documents.}}}
                \label{tab:results_gen_json}
            \end{table}

        \begin{table}[ht]
            \begin{adjustbox}{center}
                \begin{tabular}{|c|c|c|c|c|}
                    \hline
                    Model Name & GriTS$_{top}$ & GriTS$_{con}$ & TEDS & Lev-TED \\
                    \hline
                    Mistral-7B (\lift) & 0.79 & 0.86 & 0.96 & 0.96 \\
                    \hline
                    Mistral-24B (\lift) & 0.82 & 0.89 & 0.97 & 0.97 \\
                    \hline
                    Phi-4 (\lift) & 0.77 & 0.84 & 0.94 & 0.94 \\
                    \hline
                    Qwen-3 (\lift) & 0.73 & 0.78 & 0.90 & 0.92 \\
                    \hline
                    Llama-1B (\lift) & 0.68 & 0.71 & 0.87 & 0.87 \\
                    \hline
                    Llama-3B (\lift) & 0.73 & 0.78 & 0.92 & 0.93 \\
                    \hline
                    Mistral-7B (\etoe) & 0.78 & 0.83 & 0.94 & 0.94 \\
                    \hline
                    Mistral-24B (\etoe) & 0.82 & 0.88 & 0.96 & 0.97 \\
                    \hline
                    Phi-4 (\etoe) & 0.71 & 0.81 & 0.94 & 0.94 \\
                    \hline
                    Qwen-3 (\etoe) & 0.65 & 0.68 & 0.73 & 0.75 \\
                    \hline
                    Llama-1B (\etoe) & 0.70 & 0.73 & 0.87 & 0.86 \\
                    \hline
                    Llama-3B (\etoe) & 0.75 & 0.80 & 0.90 & 0.91 \\
                    \hline
                \end{tabular}
            \end{adjustbox}
            \label{tab:improvement_ratios}
            \caption{Percentage of inputs improved over single call for GPT 4o for different metrics}
        \end{table}

        \begin{table}[ht]
            \begin{adjustbox}{center}
                \begin{tabular}{|c|c|c|c|c|}
                    \hline
                    Model Name & GriTS$_{top}$ & GriTS$_{con}$ & TEDS & Lev-TED \\
                    \hline
                    Mistral-7B (\lift) & 0.25 & 0.32 & 0.25 & 377.74 \\
                    \hline
                    Mistral-24B (\lift) & 0.27 & 0.36 & 0.26 & 408.32 \\
                    \hline
                    Phi-4 (\lift) & 0.24 & 0.31 & 0.24 & 387.65 \\
                    \hline
                    Qwen-3 (\lift) & 0.20 & 0.25 & 0.21 & 349.12 \\
                    \hline
                    Llama-1B (\lift) & 0.16 & 0.18 & 0.19 & 270.28 \\
                    \hline
                    Llama-3B (\lift) & 0.20 & 0.25 & 0.22 & 342.30 \\
                    \hline
                    Mistral-7B (\etoe) & 0.24 & 0.30 & 0.24 & 371.39 \\
                    \hline
                    Mistral-24B (\etoe) & 0.26 & 0.35 & 0.26 & 412.67 \\
                    \hline
                    Phi-4 (\etoe) & 0.21 & 0.27 & 0.21 & 336.95 \\
                    \hline
                    Qwen-3 (\etoe) & 0.15 & 0.19 & 0.14 & 236.17 \\
                    \hline
                    Llama-1B (\etoe) & 0.17 & 0.20 & 0.20 & 285.96 \\
                    \hline
                    Llama-3B (\etoe) & 0.21 & 0.26 & 0.22 & 350.29 \\
                    \hline
                \end{tabular}
            \end{adjustbox}
            \label{tab:improvement_values}
            \caption{Average absolute improvement in metric against a single call to GPT 4o}
        \end{table}
        
        \begin{table}[ht]
            \begin{adjustbox}{center}
                \begin{tabular}{|c|c|c|c|c|}
                    \hline
                    Model Name & GriTS$_{top}$ & GriTS$_{con}$ & TEDS & Lev-TED \\
                    \hline
                    GPT 4o (Single Call) & [0.6672, 0.6893] & [0.5499, 0.5724] & [0.6789, 0.6949] & [465.2874, 519.4812] \\
                    \hline
                    Mistral-7B (\lift) & [0.9196, 0.9305] & [0.8713, 0.8859] & [0.9322, 0.9394] & [88.2496, 108.8308] \\
                    \hline
                    Mistral-24B (\lift) & [0.9471, 0.9551] & [0.9185, 0.9291] & [0.9481, 0.9541] & [58.9844, 73.2232] \\
                    \hline
                    Phi-4 (\lift) & [0.9090, 0.9205] & [0.8631, 0.8770] & [0.9193, 0.9269] & [94.3027, 113.8553] \\
                    \hline
                    Qwen-3 (\lift) & [0.8769, 0.8888] & [0.8078, 0.8242] & [0.8969, 0.9060] & [123.6902, 150.2994] \\
                    \hline
                    Llama-1B (\lift) & [0.8327, 0.8479] & [0.7352, 0.7547] & [0.8740, 0.8845] & [195.4633, 233.1382] \\
                    \hline
                    Llama-3B (\lift) & [0.8715, 0.8846] & [0.8015, 0.8189] & [0.8986, 0.9077] & [134.8696, 161.7537] \\
                    \hline
                    Mistral-7B (\etoe) & [0.9117, 0.9224] & [0.8585, 0.8725] & [0.9227, 0.9303] & [102.5411, 123.4352] \\
                    \hline
                    Mistral-24B (\etoe) & [0.9379, 0.9468] & [0.9052, 0.9168] & [0.9413, 0.9480] & [64.0791, 77.9916] \\
                    \hline
                    Phi-4 (\etoe) & [0.8805, 0.8937] & [0.8239, 0.8416] & [0.8910, 0.9017] & [137.9231, 168.1007] \\
                    \hline
                    Qwen-3 (\etoe) & [0.8697, 0.8879] & [0.7786, 0.8017] & [0.8448, 0.8587] & [145.3816, 176.4601] \\
                    \hline
                    Llama-1B (\etoe) & [0.8479, 0.8623] & [0.7533, 0.7722] & [0.8822, 0.8918] & [175.3041, 209.9542] \\
                    \hline
                    Llama-3B (\etoe) & [0.8819, 0.8942] & [0.8127, 0.8296] & [0.9013, 0.9097] & [122.2198, 143.7246] \\
                    \hline
                \end{tabular}
            \end{adjustbox}
            \label{tab:bootstrap_CIs}
            \caption{Bootstrap Confidence Intervals calculated at 95\% with 1000 bootstrap samples, computed for all the metrics of interest}
        \end{table}

\section{More Workflow Examples}
\begin{figure}
\centering
    \makebox[\textwidth][c]{
    \begin{tikzpicture}[
        % Define node distance: vertical then horizontal
        node distance=1.5cm and 1.5cm,
        % Style for image containers
        img_node/.style={
            inner sep=0pt, % Removes space between image and border
            draw,          % Adds a border around the image
            line width=1pt,
            rounded corners=2pt
        },
        arrow/.style={
            -{Stealth[scale=1]},
            line width=2pt,
            shorten >= 12pt,
            shorten <= 12pt
        }
    ]

    % Place the nodes with images inside the braces {}
    % Note: Change the filenames to your actual local image files
    \node [img_node] (a) {\includegraphics[width=0.5\textwidth]{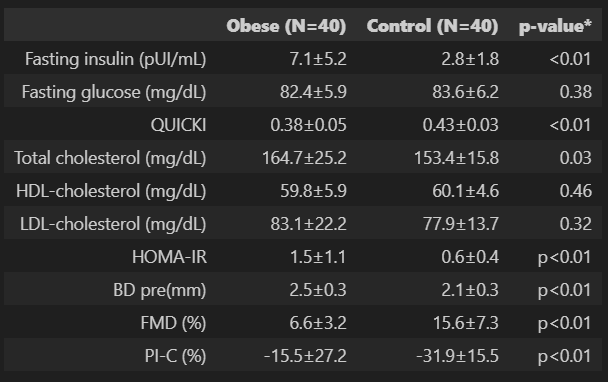}};
    
    \node [img_node, right=of a] (b) {\includegraphics[width=0.5\textwidth]{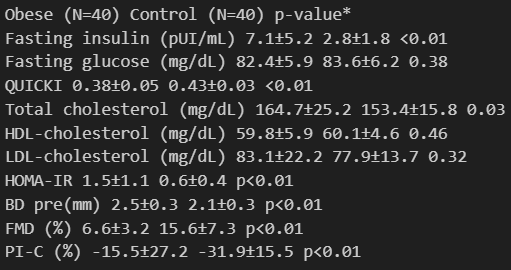}};
    
    \node [img_node, below=of b] (c) {\includegraphics[width=0.5\textwidth]{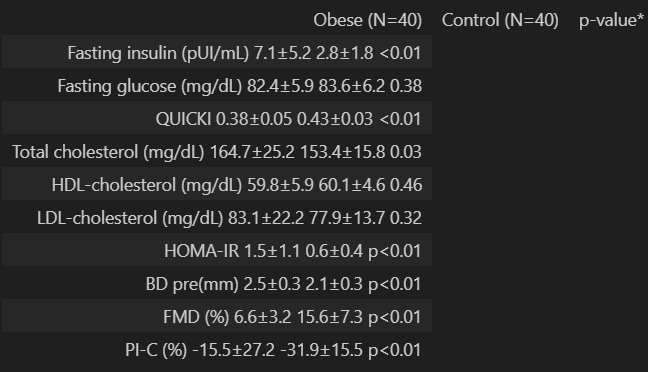}};
    
    \node [img_node, left=of c] (d) {\includegraphics[width=0.5\textwidth]{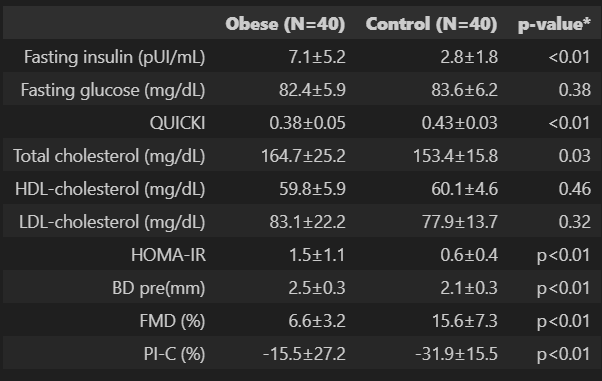}};
    
    % Draw the arrows connecting the node names (a, b, c, d)
    \draw [arrow] (a) -- (b);
    \draw [arrow] (b) -- (c);
    \draw [arrow] (c) -- (d);
    
    \end{tikzpicture}}
    \caption{Workflow visualization using images.}\label{fig:overview2}
\end{figure}

\begin{figure}
\centering
    \makebox[\textwidth][c]
    {\begin{tikzpicture}[
        % Define node distance: vertical then horizontal
        node distance=1.5cm and 1.5cm,
        % Style for image containers
        img_node/.style={
            inner sep=0pt, % Removes space between image and border
            draw,          % Adds a border around the image
            line width=1pt,
            rounded corners=2pt
        },
        arrow/.style={
            -{Stealth[scale=1]},
            line width=2pt,
            shorten >= 12pt,
            shorten <= 12pt
        }
    ]

    % Place the nodes with images inside the braces {}
    % Note: Change the filenames to your actual local image files
    \node [img_node] (a) {\includegraphics[width=0.5\textwidth]{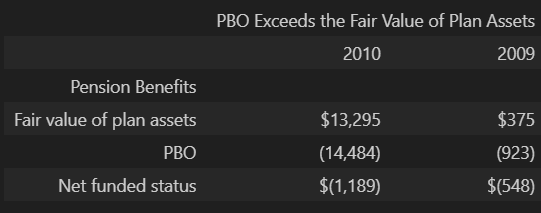}};
    
    \node [img_node, right=of a] (b) {\includegraphics[width=0.5\textwidth]{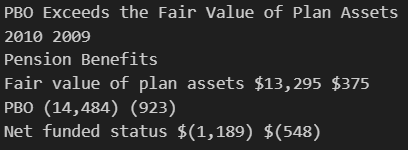}};
    
    \node [img_node, below=of b] (c) {\includegraphics[width=0.5\textwidth]{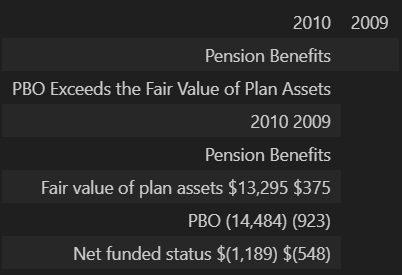}};
    
    \node [img_node, left=of c] (d) {\includegraphics[width=0.5\textwidth]{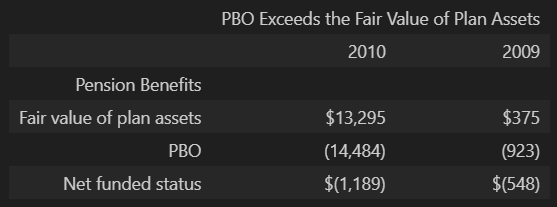}};
    
    % Draw the arrows connecting the node names (a, b, c, d)
    \draw [arrow] (a) -- (b);
    \draw [arrow] (b) -- (c);
    \draw [arrow] (c) -- (d);
    
    \end{tikzpicture}}
    \caption{Workflow visualization using images.}\label{fig:overview3}
\end{figure}

\begin{figure}
\centering
    \makebox[\textwidth][c]
    {\begin{tikzpicture}[
        % Define node distance: vertical then horizontal
        node distance=1.5cm and 1.5cm,
        % Style for image containers
        img_node/.style={
            inner sep=0pt, % Removes space between image and border
            draw,          % Adds a border around the image
            line width=1pt,
            rounded corners=2pt
        },
        arrow/.style={
            -{Stealth[scale=1]},
            line width=2pt,
            shorten >= 12pt,
            shorten <= 12pt
        }
    ]

    % Place the nodes with images inside the braces {}
    % Note: Change the filenames to your actual local image files
    \node [img_node] (a) {\includegraphics[width=0.5\textwidth]{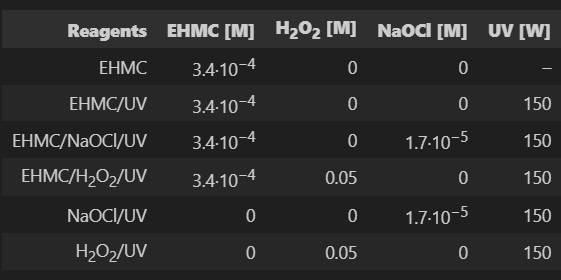}};
    
    \node [img_node, right=of a] (b) {\includegraphics[width=0.5\textwidth]{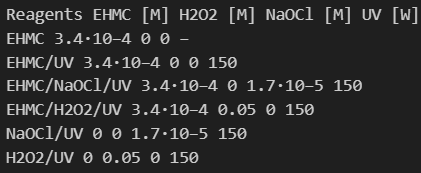}};
    
    \node [img_node, below=of b] (c) {\includegraphics[width=0.5\textwidth]{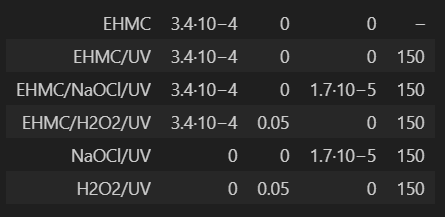}};
    
    \node [img_node, left=of c] (d) {\includegraphics[width=0.5\textwidth]{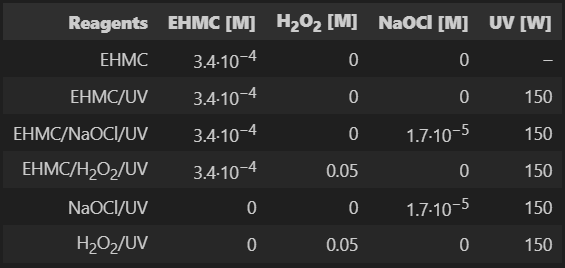}};
    
    % Draw the arrows connecting the node names (a, b, c, d)
    \draw [arrow] (a) -- (b);
    \draw [arrow] (b) -- (c);
    \draw [arrow] (c) -- (d);
    
    \end{tikzpicture}
    }\caption{Workflow visualization using images.}\label{fig:overview4}
\end{figure}

\begin{figure}
\centering
    \makebox[\textwidth][c]{
    \begin{tikzpicture}[
        % Define node distance: vertical then horizontal
        node distance=1.5cm and 1.5cm,
        % Style for image containers
        img_node/.style={
            inner sep=0pt, % Removes space between image and border
            draw,          % Adds a border around the image
            line width=1pt,
            rounded corners=2pt
        },
        arrow/.style={
            -{Stealth[scale=1]},
            line width=2pt,
            shorten >= 12pt,
            shorten <= 12pt
        }
    ]

    % Place the nodes with images inside the braces {}
    % Note: Change the filenames to your actual local image files
    \node [img_node] (a) {\includegraphics[width=0.5\textwidth]{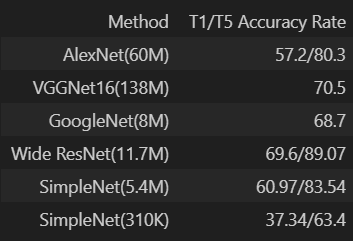}};
    
    \node [img_node, right=of a] (b) {\includegraphics[width=0.5\textwidth]{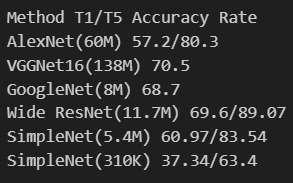}};
    
    \node [img_node, below=of b] (c) {\includegraphics[width=0.5\textwidth]{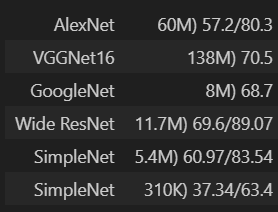}};
    
    \node [img_node, left=of c] (d) {\includegraphics[width=0.5\textwidth]{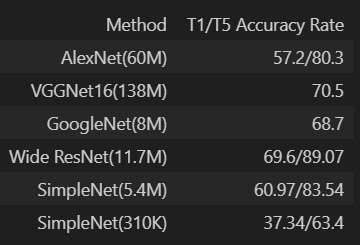}};
    
    % Draw the arrows connecting the node names (a, b, c, d)
    \draw [arrow] (a) -- (b);
    \draw [arrow] (b) -- (c);
    \draw [arrow] (c) -- (d);
    
    \end{tikzpicture}}
    \caption{Workflow visualization using images.}\label{fig:overview5}
\end{figure}

\begin{figure}[hbtp]
    \centering
    \makebox[\textwidth][c]
    {\begin{tikzpicture}[
        % Define node distance: vertical then horizontal
        node distance=1.5cm and 1.5cm,
        % Style for image containers
        img_node/.style={
            inner sep=0pt, % Removes space between image and border
            draw,          % Adds a border around the image
            line width=1pt,
            rounded corners=2pt
        },
        arrow/.style={
            -{Stealth[scale=1]},
            line width=2pt,
            shorten >= 12pt,
            shorten <= 12pt
        }
    ]

    % Place the nodes with images inside the braces {}
    % Note: Change the filenames to your actual local image files
    \node [img_node] (a) {\includegraphics[width=0.5\textwidth]{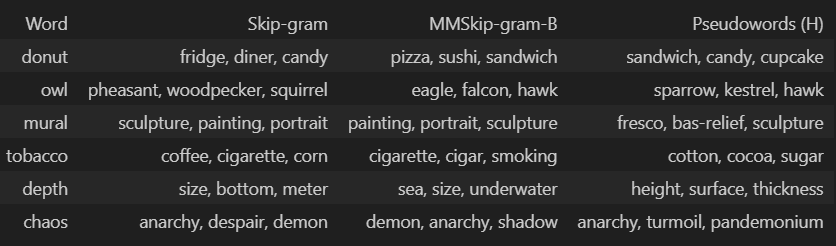}};
    
    \node [img_node, right=of a] (b) {\includegraphics[width=0.5\textwidth]{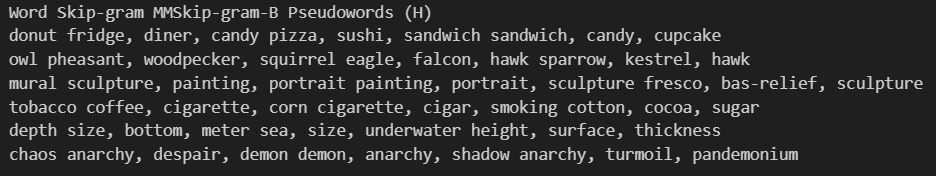}};
    
    \node [img_node, below=of b] (c) {\includegraphics[width=0.5\textwidth]{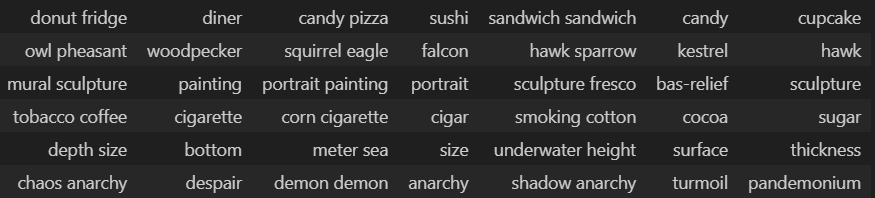}};
    
    \node [img_node, left=of c] (d) {\includegraphics[width=0.5\textwidth]{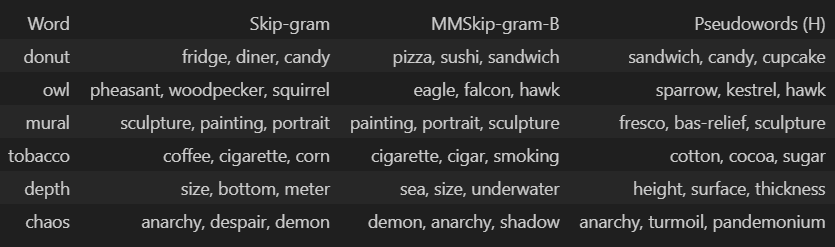}};
    
    % Draw the arrows connecting the node names (a, b, c, d)
    \draw [arrow] (a) -- (b);
    \draw [arrow] (b) -- (c);
    \draw [arrow] (c) -- (d);
    
    \end{tikzpicture}
    }\caption{Workflow visualization using images.}\label{fig:overview6}
\end{figure}

Following the style of the illustration presented in Figure~\ref{fig:overview}, we present a few
other illustrations of last-mile repair using fine-tuned SLMs on the table explicitation problem
in 
Figure~\ref{fig:overview2},
Figure~\ref{fig:overview3},
Figure~\ref{fig:overview4},
Figure~\ref{fig:overview5},
and
Figure~\ref{fig:overview6}.

\ignore{
\begin{figure}
\centering
    \makebox[\textwidth][c]{\begin{tikzpicture}[
        % Define node distance: vertical then horizontal
        node distance=1.5cm and 1.5cm,
        % Style for image containers
        img_node/.style={
            inner sep=0pt, % Removes space between image and border
            draw,          % Adds a border around the image
            line width=1pt,
            rounded corners=2pt
        },
        arrow/.style={
            -{Stealth[scale=1]},
            line width=2pt,
            shorten >= 12pt,
            shorten <= 12pt
        }
    ]

    % Place the nodes with images inside the braces {}
    % Note: Change the filenames to your actual local image files
    \node [img_node] (a) {\includegraphics[width=0.5\textwidth]{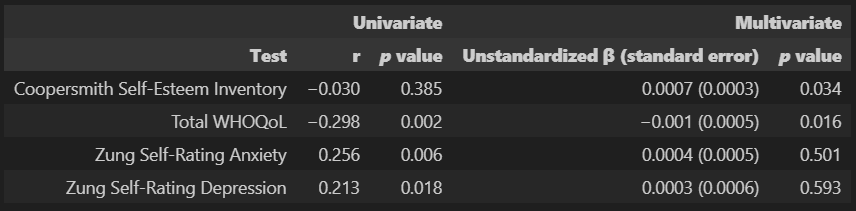}};
    
    \node [img_node, right=of a] (b) {\includegraphics[width=0.5\textwidth]{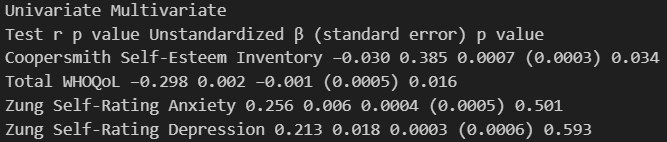}};
    
    \node [img_node, below=of b] (c) {\includegraphics[width=0.5\textwidth]{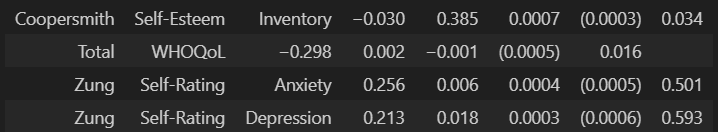}};
    
    \node [img_node, left=of c] (d) {\includegraphics[width=0.5\textwidth]{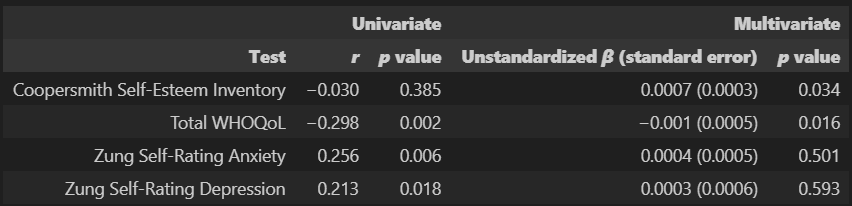}};
    
    % Draw the arrows connecting the node names (a, b, c, d)
    \draw [arrow] (a) -- (b);
    \draw [arrow] (b) -- (c);
    \draw [arrow] (c) -- (d);
    
    \end{tikzpicture}}
    \caption{Workflow visualization using images.}

\end{figure}
\endignore}

\section{Prompts, Hyperparameters, Dataset Splits} 
\label{app:prompts}
% TODO: Figure out problem with encoding and double quotes
    The models were all fine-tuned for 5 epochs, with an effective batch size of 128. 50 warmup steps up to a learning rate of $5\times 10^{-5}$ were used. The model was evaluated on the validation set after every epoch, and at the end of the training run the checkpoint with the smallest validation loss was chosen. Training was done at half-precision using the bfloat16 format and an 8-bit version of the Adam optimizer. LORA was used with rank 16 and $\alpha = 32$.

    \begin{table}[t]
        \begin{center}
        \begin{tabular}{llll}
        \toprule
        \multicolumn{1}{c}{\bf Dataset} &\multicolumn{1}{c}{\bf Train}  &\multicolumn{1}{c}{\bf Val} &\multicolumn{1}{c}{\bf Test} \\
        \midrule
        PubTabNet         & 3464 & 422 & 877 \\
        FinTabNet         & 2714 & 361 & 805 \\
        SciTSR            & 2789 & 350 & 914 \\
        \bottomrule
        \end{tabular}
        \end{center}
        \caption{Dataset Contribution to Splits}
        \label{tab:dataset_split}
    \end{table}

    We have two prompts, one for table explicitation and one for table repair.
    
    System Prompt(Repair):

    \begin{lstlisting}
You are an expert in interpreting various table formats. Your task is to generate fixed HTML tables from unstructured text, and optionally a table that may or may not contain errors. 
    \end{lstlisting}
%Your primary goal is to **generalize** the process of table extraction

    User Prompt(Repair):

    \begin{lstlisting}
I have an unstructured text representation of a table and an html representation of the same table.
Instructions:
-Analyze the given ''Raw Input **unstructured** Text'' and html table to accurately identify rows, columns, and data cells.
-Ensure that the output table maintains the same row and column structure as the given ''Raw Input Text''.
-Keep the table rows and columns in the same order and structure as they appear in the given ''Raw Input Text''.
-Include all content from the current given ''Raw Input Text'' without omission.
-Do not add any content in the fixed HTML table if it is not present in the given ''Raw Input Text''. 
-Ensure that the fixed HTML table is well-formed and valid, and enclosed inside <table></table> tags.
''Raw Input Text'':
{{Input Text Here}}

''HTML Table'':
{{Broken HTML Table Here}}
    \end{lstlisting}

    System Prompt(Explicitation):

    \begin{lstlisting}
You are an expert in interpreting various table formats. Your task is to generate fixed HTML tables from unstructured text. 
    \end{lstlisting}
    
    User Prompt(Explicitation):
    \begin{lstlisting}
I have an unstructured text representation of a table.
Instructions:
-Analyze the given ''Raw Input **unstructured** Text'' and html table to accurately identify rows, columns, and data cells.
-Ensure that the output table maintains the same row and column structure as the given ''Raw Input Text''.
-Keep the table rows and columns in the same order and structure as they appear in the given ''Raw Input Text''.
-Include all content from the current given ''Raw Input Text'' without omission.
-Do not add any content in the fixed HTML table if it is not present in the given ''Raw Input Text''. 
-Ensure that the fixed HTML table is well-formed and valid, and enclosed inside <table></table> tags.
''Raw Input Text'':
{{Input Text Here}}
    \end{lstlisting}

\end{document}